\documentclass[10pt, a4paper]{article}
\usepackage{lrec}
\usepackage{multibib}
\newcites{languageresource}{Language Resources}
\usepackage{graphicx}
\usepackage{tabularx}
\usepackage{soul}
\usepackage{qtree}
\usepackage{multirow}
\usepackage{arabtex}
\usepackage[T1]{fontenc}

\usepackage{epstopdf}
\usepackage[latin1]{inputenc}

\usepackage{url}
\usepackage{listings}
\usepackage{color}
\definecolor{gray}{rgb}{0.4,0.4,0.4}
\definecolor{darkblue}{rgb}{0.0,0.0,0.6}
\definecolor{cyan}{rgb}{0.0,0.6,0.6}

\newcommand{\hide}[1]{}

\newcommand{\AMADDA}{{\={A}}}
\newcommand{\AHAMZAUP}{{\^{A}}}

\newcommand{\AHAMZADN}{{\v{A}}}
\newcommand{\YHAMZA}{{\^{y}}}
\newcommand{\TAMARBUTA}{{$\hbar$}}

\newcommand{\SHIN}{{\v{s}}}
\newcommand{\DAD}{{\v{D}}} %Z
\newcommand{\AYN}{{$\varsigma$}}
\newcommand{\GAYN}{{$\gamma$}}

\newcommand{\FATHATAN}{{\~{a}}}

\newcommand{\SHADDA}{{$\sim$}}

\newcommand{\CTB} {CATiB}

\title{\textbf{An Arabic Dependency Treebank in the Travel Domain}}

\name{Dima Taji, Jamila El Gizuli, Nizar Habash}

\address{Computational Approaches to Modeling Language Lab\\ New York University Abu Dhabi, UAE \\
        \{dima.taji,nizar.habash\}@nyu.edu\\}

\abstract{
In this paper we present a dependency treebank of travel domain sentences in Modern Standard Arabic. 
The text comes from a translation of the English equivalent sentences in the Basic Traveling Expressions Corpus.
The treebank dependency representation is in the style of the Columbia Arabic Treebank. The paper motivates the effort
and discusses the construction process and guidelines. We also present parsing results and discuss the effect of domain
and genre difference on parsing.
\\ \newline  }

\begin{document}

\maketitleabstract
\setarab
\novocalize

\section{Introduction}

Treebanks, or annotated corpora, are essential for Natural Language Process (NLP) tasks.  Such tasks include building lexicons, inferencing grammars, and creating computational analyzers, which can all be improved by creating treebanks with different kinds of linguistic annotations \cite{abeille2012treebanks}. %Taken from introduction - need to specify?
Treebanks with rich annotation and good quality are very expensive resources to create. %cite? who?
They require a large number of man-hours to create and audit. %audit?

Treebanks can be in multiple genres, or genre-specific.\footnote{The terms {\it domain}, {\it genre}, {\it topic} and {\it style} have been discussed a lot in the field \cite{lee2002genres,van2015s,ide2017handbook}, and many authors discussed their ambiguous and overlapping use. For the rest of this paper we use the term travel domain, following \newcite{takezawa2007multilingual} whose corpus was the basis for the translated corpus we treebank.} %generally? usually? mostly? what's the proof?
However, there is a tradeoff between the cost of the size, the diversity of a corpus, and having enough content in one genre or domain to be able to make generalizations. As a result, many treebanks tend to be predominantly of one specific genre, but may add some samples of other genres. For example, the Hindi/Urdu Treebank \cite{bhat2017hindi} is predominantly in the news domain with 85.3\% of its sentences coming from news articles, and only 14.7\% from other domains (9.7\% from conversations, and 5\% from the travel domain). \newcite{webber2009genre} shows that the Penn Treebank \cite{Marcus:1994} consists of 90.1\% news articles, 4.9\% essays, 2.6\% summaries, and 2.4\% letters, and it is still considered to be a news domain treebank. Similarly, \newcite{maamouri2010penn} demonstrate that the Penn Arabic Treebank (PATB) \cite{Maamouri:2004} consists of 39.9\% newswire text, 28.2\% broadcast news, 18.6\% broadcast conversation in both Standard and Dialectal Arabic, and 13.3\% web texts.

In this paper we describe a small Modern Standard Arabic (MSA) treebank, created using a travel corpus. This treebank will be the seed of a larger multi-genre, and multi-dialect Arabic treebank. %is this statement too confident?
The corpus we are using is part of an MSA translation by \newcite{eck2005overview} of the Basic Travel Expression Corpus (BTEC) \cite{takezawa2007multilingual}, henceforth MSABTEC. As far as we know, there is no treebank based on this corpus.

In Section~2, we discuss related work followed by a description of the corpus we annotate in Section~3. In Section~4, we discuss the annotation format; and in Section~5 the annotation process. Finally, we present some results on benchmarking parsing on our corpus and a  comparison with a major news-domain Arabic treebank in Section~6.

\section{Related Work}

BTEC is a collection of conversational phrases that cover various situations in the travel domain in Japanese, and their translations into English and Chinese \cite{takezawa2007multilingual}.  The sentences in the corpus were collected from bilingual travel experts, and were based from their experience rather than being transcribed. The corpus was later translated into more languages including Arabic \cite{eck2005overview}, where it was used for evaluating machine translation systems.

Another treebank that included phrases from the travel domain is the Hindi/Urdu treebank \cite{bhat2017hindi}. Even though the majority of the treebank comes from news sources, it contains 15K words, making up 1,058 sentences relating to heritage and tourism. This part of the data was specifically added to counteract the bias that could result from using data in one specific domain, news in this instance. The treebank contains dependency, phrase-structure, and PropBank-inspired \cite{kingsbury2002adding} annotations. 

The Penn Treebank is a well known resource, that contains phrases mostly from the news domain. The treebank was annotated for genres as part of the Penn Discourse Treebank \cite{MPJW:2004}, and \newcite{webber2009genre} shows that the different genres can have different characteristics.
%NYH: such as? say more if we need space.

The Penn Arabic Treebank (PATB) is the primary treebank for work on Arabic syntactic analysis. It uses a phrase-structure representation; but has been converted to other dependency formalisms \cite{habash2009catib,taji2017universal}.  The PATB contains various parts that come from different domains and resources. PATB comes in 12 parts \cite{diab2013ldc}, that are mostly from news or web sources \cite{maamouri2010penn}. Other related treebanks were developed by the Linguistic Data Consortium (LDC) in various dialects such as Egyptian \cite{maamouri2012egyptian}, and Levantine \cite{maamouri2006developing}, where the data came from transcribing recorded conversations. 

The first dependency Arabic treebank was the Prague Arabic Dependency Treebank (PADT)  \cite{NEMLAR:PADT:2004}.
 It employed a  multi-level description scheme for functional morphology, analytical dependency syntax, and tectogrammatical representation of linguistic meaning.

Another large Arabic treebank is the Columbia Arabic Treebank (CATiB) \cite{habash2009catib}. CATiB has around 250K words that were annotated directly in it in addition to the full converted PATB. CATiB focuses on news domain text in Standard Arabic.
 Most recently, \newcite{taji2017universal} converted the PATB into the formalism of the universal dependency (UD) project \cite{NIVRE16.348} via an intermediate step of mapping to CATiB dependencies.  

The Quran Corpus is another important Arabic syntactic corpus of  the very specific genre of holy scripture \cite{Dukes:2010b}. It has its own representation scheme which is a hybrid dependency and constituency. 

In this work, we annotate in the format of the CATiB treebank and compare to UD representations. And we present a comparison with the news domain as captured in the PATB.

\hide{
NYH:  Add a pointer to the Hindi/Urdu treebank... they had tourism data: 
%https://link.springer.com/content/pdf/10.1007%2F978-94-024-0881-2_24.pdf
\cite{bhat2017hindi} ... added to Extra bib

NYH: since it is possible you may have trouble getting the PDFs for the springer material. I put them in a directory under the paper directory

NYH: Add a pointer to various PATB parts (beyond 3 ... it goes high to 14 or more).
NYH:  another one for English http://www.aclweb.org/anthology/P09-1076

NYH: BTEC English paper which discusses the genre...
I added it to Extra bib: \cite{takezawa2007multilingual}
%http://citeseerx.ist.psu.edu/viewdoc/download;jsessionid=E2D412397B796CE6847BADB843FC03E2?doi=10.1.1.108.2869&rep=rep1&type=pdf
}

\section{Our Corpus}
For our corpus, we selected the MSA translation of BTEC \cite{eck2005overview}.%, henceforth MSABTEC.  
Our selection contains 2,000 sentences making a total of 
15,929 words (7.9 words/sentence). 
%NYH: are these PATB tokens or words?
%How was the 2,000 sentences selected? They are test, but according to which data division? (ADD LINK TO LREC PAPER ON MADAR)....
The sentence choice overlapped with the test set used in another project that focuses on machine translation and language identification (Anonymous, {\it under review}). % NYH - if we are 3 pages, we should not use this: \footnote{More details will be presented in the final version of the paper.}
The text of the corpus, coming from BTEC, is full of travel related expressions such as inquiring about the prices of hotel rooms, asking for directions, requesting help, ordering food, etc. 
Being conversational, it also has a high percentage of first and second person pronouns and conjugations.
Below are examples of sentences from MSABTEC:
\begin{footnotesize}
\begin{itemize}
\item <'a.htAj 'ilY .tbyb.> \emph{{\AHAMZAUP}HtAj {\AHAMZADN}lY Tbyb.}\footnote{Arabic 
transliteration is presented in the
  Habash-Soudi-Buckwalter scheme \cite{HSB-TRANS:2007}.}
%(in alphabetical order)\\
%\addtolength{\tabcolsep}{-5.3pt}
%\begin{tabular}{cccccccccccccccccccccccccccc}
%<'a> & <b> & <t> & <_t> & <^g> & <.h> & <_h> & <d> & <_d> & <r> & <z> & <s> & <^s> & <.s> & <.d> & <.t> & <.z> & <`> & <.g> & <f> & <q> & <k> & <l> & <m> & <n> & <h> & <w> & <y> \\
%{\AHAMZAUP} & b & t & {\THA} & j & H & x & d & {\DHA} & r & z & s & {\SHIN} & S & D & T & {\DAD} & {\AYN} & {\GAYN} & f & q & k & l & m & n & h & w & y\\
%\end{tabular}
%\\
%and the additional symbols: '~<"'>, {\AHAMZAUP}~<'a>,
%{\AHAMZADN}~<'i>, {\AMADDA}~<'A>, {\WHAMZA}~<|u'u>, {\YHAMZA}~<|Y'>,
%{\TAMARBUTA}~<T>, {\AMAQSURA}~<Y>. }\\ 
     	`I need a doctor.'
\item <krymT wskr?> \emph{krym{\TAMARBUTA} wskr?}
	`Cream and sugar?'
	
%Add an example with question word.
\item <'ayn 'aqrb m.hl jzArT?>  \emph{{\AHAMZAUP}yn {\AHAMZAUP}qrb mHl jzAr{\TAMARBUTA}?}

\verb| |
`Where is the nearest butcher?'

%\item <'ayn AlhAtif?>  \emph{{\AHAMZAUP}yn AlhAtf?}
%`Where is the phone?'

\end{itemize}
\end{footnotesize}

%\hide{
%Table~\ref{example-sentences} shows two examples of sentences from BTEC, and their translations in MSABTEC. 
%
%\begin{table}[!h]
%\begin{center}
%\setlength{\tabcolsep}{3pt}
%\begin{tabular}{l}
%\hline
%	<'a.htAj 'ilY .tbyb>\\
%	{\AHAMZAUP}HtAj {\AHAMZADN}lY Tbyb\footnote{
%All Arabic transliterations are provided in the
%Habash-Soudi-Buckwalter transliteration scheme
%\cite{HSB-TRANS:2007}. This scheme extends  Buckwalter's
%transliteration scheme \cite{Buckwalter:2002} to increase its
%readability while maintaining the 1-to-1 correspondence with Arabic
%orthography as represented in standard encodings of Arabic, i.e.,
%Unicode, CP-1256, etc.  The following are the only differences from
% Buckwalter's scheme (which is indicated in parentheses):
%{\AMADDA}~<'A>~($|$),
%{\AHAMZAUP}~<'a>~($>$),
%{\WHAMZA}~<|u'u>~(\&),
%{\AHAMZADN}~<'i>~($<$),   
%{\YHAMZA}~ \setsindhi<|'y>\setarab~($\}$), 
%{\TAMARBUTA}~<T>~(p),
%{\THA}~<_t>~(v),
%{\DHA}~<_d>~($\ast$),
%{\SHIN}~<^s>~(\$),
%{\DAD}~<.z>~(Z),
%{\AYN}~<`>~(E),
%{\GAYN}~<.g>~(g),
%{\AMAQSURA}~<Y>~(Y),
%{\FATHATAN}~<|--"aN>~(F),
%{\DAMMATAN}~<|--"uN>~(N),
%{\KASRATAN}~<|--"iN>~(K),
%{\DAGGER}~<|B"_a>~(`).}\\ 
%     	I need a doctor\\
%\hline
%	<krymT wskr ?>\\
%	krym{\TAMARBUTA} wskr ?\\
%	Cream and sugar ?\\
%\hline
%  \end{tabular}
%\caption{Example Sentences from MSABTEC}
%\label{example-setences}
%\end{center}
%\end{table}
%}

%Insert a couple of example sentences
%a question, and a non-question

\section{Annotation Format}
To maximize compatibility with previous efforts, we followed the Columbia Arabic Treebank (CATiB) \cite{habash2009catib} annotation guidelines and tokenization schemes used by previous Arabic treebanks. We chose this format because it uses traditional Arabic grammar as the inspiration for its relational labels and dependency structure \cite{habash2009catib}, making it intuitive for Arabic speakers, and allowing for faster annotation. In addition, this format can be automatically enriched with more morphological features \cite{alkuhlani-habash-roth:2013:NAACL-HLT}, and converted into other dependency formats such as the Universal Dependency format \cite{taji2017universal}.  Except for a number of minor specifications for some new syntactic constructions, there was no change to the guidelines for tokenization, part-of-speech (POS) tag set and relations.

\subsection{Tokenization} The tokenization followed in the treebank creation is the same tokenization scheme used in PATB. This scheme tokenizes all the clitics, except for the definite article +<Al> {\em Al+} `the' \cite{pasha2014madamira}. The 2,000 sentences in our corpus consist of 18,628 tokens (manually checked). %do we include the number of tokens here or under the Our Corpus section? Would it make sense to include the number of untokenized words there, and the number of tokens here, or is that just going to be confusing?

\subsection{Annotation Scheme} For our treebank, we followed the CATiB dependency annotation scheme. This scheme is designed to be speedy for annotation, and intuitive for Arabic speakers. We also used the guidelines that were prepared for the CATiB annotation project \cite{CATIB:MEDAR:2009}. 
%is there another citation specifically for the guidelines? NYH:No

\subsubsection{POS Tags} The CATiB annotation scheme uses six  POS tags which are \textbf{NOM} for all nominals excluding proper nouns; \textbf{PROP} for proper nouns; \textbf{VRB} for active-voice verbs; \textbf{VRB-PASS} for passive-voice verbs; \textbf{PRT} for particles, which include prepositions and conjunctions; and \textbf{PNX} for punctuation marks. 

\subsubsection{Relations} There are eight relations used in the CATiB scheme: \textbf{SBJ} for the subjects of verbs and the topics of simple nominal sentences; \textbf{OBJ} for the objects of verbs, prepositions, or deverbal nouns; \textbf{TPC} for the topics of complex nominal sentences which contain  explicit pronominal referents; \textbf{PRD} for the complements of the extended copular constructions; \textbf{IDF} for marking the possessive nominal construction ({\it idafa}); \textbf{TMZ} for marking  the {\it specification} nominal construction ({\it tamyiz}); \textbf{MOD} for general modification of  verbs or nominals; and, finally,  \textbf{---} for marking flat constructions such as first-last proper name sequences. 

%does this look too much like the original text in the CATiB paper?

%NYH: I added this... there may be more to say based on what Jamila did:
\subsubsection{Syntactic Structures} 
\begin{figure}[htp]
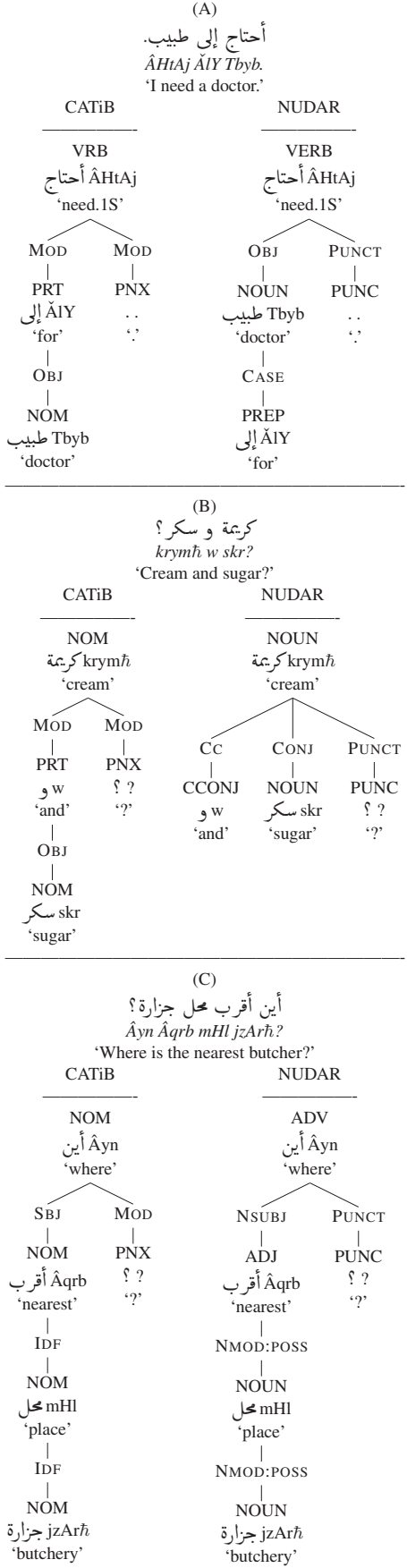

%{\scriptsize

\begin{footnotesize}
%{\tiny
\begin{center}
\scalebox{0.83}{
 \begin{tabular}{cc}
%{\normalsize
%\multicolumn{2}{c}{(A) {<'a.htAj 'ilY .tbyb.>} 
% {\emph{{\AHAMZAUP}HtAj {\AHAMZADN}lY Tbyb.}}
% {`I need a doctor.'} } \\
\multicolumn{2}{c}{(A)}\\ 
\multicolumn{2}{c}{<'a.htAj 'ilY .tbyb.>} \\ 
 \multicolumn{2}{c}{\emph{{\AHAMZAUP}HtAj {\AHAMZADN}lY Tbyb.}}\\ 
 \multicolumn{2}{c}{`I need a doctor.'} \\ 
 
{\label{doctor-catib}\Tree [.{\CTB}\\----------------\\VRB\\{<'a.htAj>  {{\AHAMZAUP}HtAj}}\\{`need.1S'} 
         		[.{\sc Mod} [.PRT\\{<'ilY> {\AHAMZADN}lY}\\{`for'} 
				[.{\sc Obj} [.NOM\\{<.tbyb> Tbyb}\\{`doctor'} ] ] ] ]
         		[.{\sc Mod} [.PNX\\{<.> .}\\{`.'} ] ] ]}
&
{\label{doctor-ud}\Tree [.NUDAR\\----------------\\VERB\\{<'a.htAj>  {{\AHAMZAUP}HtAj}}\\{`need.1S'} 
         		[.{\sc Obj} [.NOUN\\{<.tbyb> Tbyb}\\{`doctor'}
				[.{\sc Case} [.PREP\\{<'ilY> {\AHAMZADN}lY}\\{`for'} ] ] ] ]
         		[.{\sc Punct} [.PUNC\\{<.> .}\\{`.'} ] ] ]}
 \\
%\\\hline
 \multicolumn{2}{c}{-------------------------------------------------------------------}\\
%\multicolumn{2}{l}{
% (B) {<krymT w skr?>}
%  {\emph{krym{\TAMARBUTA} w skr?}} 
%  {`Cream and sugar?'}}\\
\multicolumn{2}{c}{(B)}\\
\multicolumn{2}{c}{<krymT w skr?>}\\
\multicolumn{2}{c}{\emph{krym{\TAMARBUTA} w skr?}}\\
\multicolumn{2}{c}{`Cream and sugar?'}\\

{\label{cream-catib}\Tree [.{\CTB}\\----------------\\NOM\\{<krymT>  {krym{\TAMARBUTA}}}\\{`cream'} 
         		[.{\sc Mod} [.PRT\\{<w> w}\\{`and'} 
				[.{\sc Obj} [.NOM\\{<skr> skr}\\{`sugar'} ] ] ] ]
         		[.{\sc Mod} [.PNX\\{<?> ?}\\{`?'} ] ] ]}
	&
{\label{cream-ud}\Tree [.NUDAR\\----------------\\NOUN\\{<krymT>  {krym{\TAMARBUTA}}}\\{`cream'} 
         		[.{\sc Cc} [.CCONJ\\{<w> w}\\{`and'} ] ]
			[.{\sc Conj} [.NOUN\\{<skr> skr}\\{`sugar'} ] ] 
         		[.{\sc Punct} [.PUNC\\{<?> ?}\\{`?'} ] ] ]}
 \\
  \multicolumn{2}{c}{-------------------------------------------------------------------}\\
%\\\hline
%\multicolumn{2}{l}{
% (C) {<'ayn 'aqrb m.hl jzArT?>}
% {\emph{{\AHAMZAUP}yn {\AHAMZAUP}qrb mHl jzAr{\TAMARBUTA}?}}
% {`Where is the nearest butcher?'} }\\

\multicolumn{2}{c}{(C)}\\
\multicolumn{2}{c}{<'ayn 'aqrb m.hl jzArT?>}\\
 \multicolumn{2}{c}{\emph{{\AHAMZAUP}yn {\AHAMZAUP}qrb mHl jzAr{\TAMARBUTA}?}}\\
\multicolumn{2}{c}{`Where is the nearest butcher?'}\\

{\label{question-catib}\Tree [.{\CTB}\\----------------\\NOM\\{<'ayn> {\AHAMZAUP}yn}\\{`where'} 
         		[.{\sc Sbj} [.NOM\\{<'aqrb>  {{\AHAMZAUP}qrb}}\\{`nearest'} 
				[.{\sc Idf} [.NOM\\{<m.hl> mHl}\\{`place'}
					[.{\sc Idf} [.NOM\\{<jzArT> jzAr{\TAMARBUTA}}\\{`butchery'} ] ] ] ] ] ]
         		[.{\sc Mod} [.PNX\\{<?> ?}\\{`?'} ] ] ]}
	&
{\label{question-ud}\Tree [.NUDAR\\----------------\\ADV\\{<'ayn> {\AHAMZAUP}yn}\\{`where'} 
         		[.{\sc Nsubj} [.ADJ\\{<'aqrb>  {{\AHAMZAUP}qrb}}\\{`nearest'} 
			[.{\sc Nmod:poss} [.NOUN\\{<m.hl> mHl}\\{`place'}
				[.{\sc Nmod:poss} [.NOUN\\{<jzArT> jzAr{\TAMARBUTA}}\\{`butchery'} ] ] ] ] ] ]
         		[.{\sc Punct} [.PUNC\\{<?> ?}\\{`?'} ] ] ]}
 \end{tabular}
 }
 \end{center}
 \end{footnotesize}

% \multicolumn{2}{c}{-------------------------------------------------------------------}\\
%%\\\hline
%%\multicolumn{2}{l}{
%% (C) {<'ayn 'aqrb m.hl jzArT?>}
%% {\emph{{\AHAMZAUP}yn {\AHAMZAUP}qrb mHl jzAr{\TAMARBUTA}?}}
%% {`Where is the nearest butcher?'} }\\
%
%\multicolumn{2}{c}{(C)}\\
%\multicolumn{2}{c}{<'ayn 'aqrb m.hl jzArT?>}\\
% \multicolumn{2}{c}{\emph{{\AHAMZAUP}yn {\AHAMZAUP}qrb mHl jzAr{\TAMARBUTA}?}}\\
%\multicolumn{2}{c}{`Where is the nearest butcher?'}\\
%
%
%{\label{question-catib}\Tree [.{\CTB}\\----------------\\NOM\\{<'aqrb>  {{\AHAMZAUP}qrb}}\\{`nearest'} 
%         		[.{\sc Mod} [.NOM\\{<'ayn> {\AHAMZAUP}yn}\\{`where'} ] ]
%			[.{\sc Idf} [.NOM\\{<m.hl> mHl}\\{`place'}
%				[.{\sc Idf} [.NOM\\{<jzArT> jzAr{\TAMARBUTA}}\\{`butchery'} ] ] ] ]
%         		[.{\sc Mod} [.PNX\\{<?> ?}\\{`?'} ] ] ]}
%	&
%{\label{question-ud}\Tree [.NUDAR\\----------------\\ADJ\\{<'aqrb>  {{\AHAMZAUP}qrb}}\\{`nearest'} 
%         		[.{\sc Advmod} [.ADV\\{<'ayn> {\AHAMZAUP}yn}\\{`where'} ] ]
%			[.{\sc Nmod:poss} [.NOUN\\{<m.hl> mHl}\\{`place'}
%				[.{\sc Nmod:poss} [.NOUN\\{<jzArT> jzAr{\TAMARBUTA}}\\{`butchery'} ] ] ] ]
%         		[.{\sc Punct} [.PUNC\\{<?> ?}\\{`?'} ] ] ]}
% \end{tabular}
% }
% \end{center}
% \end{footnotesize}

\caption{The structures for example trees from the MSABTEC Treebank in CATiB format, and their counterpart in the Arabic Universal Dependency (NUDAR) format \cite{taji2017universal}.}
\label{trees}
\end{figure}

Since the original CATiB treebank, as with the Penn Arabic treebank, was focused on the news genre, there were many syntactic constructions that MSABTEC introduced that needed special attention. In particular, there was an abundance of interrogatives, and first and second person statements in MSABTEC compared to CATiB. To address these constructions, additional guideline specifics and clarifications were added. All of these extensions followed naturally from the spirit of the original guidelines. 
 For example, an interrogative pronoun such as <mn> {\it man} `who/whom' is often sentence-initial, but it can be the subject or the object of a verb:  <mn sm` +k?> {\it man sami{\AYN}a +ka?} `who heard you?' versus
<mn sm`t?> {\it man sami{\AYN}ta?} `whom did you hear?'.
Similarly, in Figure~\ref{trees}~(C), the interrogative adverb <'ayn> \emph{{\AHAMZAUP}yn} `where' is treated as the predicate head of a copular sentence since that is the syntactic role of the answer to the question.
For another common example   in this genre, single word interjections such as <'Asif> {\it {\AMADDA}sf} `sorry' or \vocalize <aN><^skr> \novocalize  {\it {\SHIN}krA{\FATHATAN}} `thanks' are treated as independent sentence trees that attached directly to the main root of the sentence they appear in.

% 
%1- Interrogatives when used as a sentence initial (as in a question) are given the relational tag of their position in the answer to the question. Section 4.2.3 in the paper addresses that but no slide has been added to the guide for it. However, I wanted to point out the example used in the paper ??? ???? ??? ?????. Shouldn't ??? serve as the head of this sentence since it's a PRD? That was the practice in the last annotation task. 
%
%2- Subordinating conjunctions like ????? ?????? ?????  and any others that were segmented apart in my parsing task, are not addressed in the guide besides slide 47 which addresses SC in general and makes mention of segmented ones but not how to annotate them and what follows them.
%
%3- Since the corpus contained travel expressions, there were many sentences that had ????? ???????. It would be useful to add a slide addressing ???? (as followed by a SUB with the attached pronoun being an OBJ) and ???? (as ???? ???????) since it is often confusing to others.
%
%4- Lastly, in cases of interjections like "sorry" and "please" as sentences on their own, we agreed to attach separately to the head node on their own. If this still is the case, then a slide should be added to demonstrate that expressions comprising of more than one sentence should all be separately attached to the head node since the old guide doesn't contain that.
%

\subsection{Interface} The annotation was done using the TrEd annotation interface \cite{pajas-tred:2008}, which was also used by \newcite{habash2009catib} for CATiB annotation.

Figure~\ref{trees} illustrates the annotation scheme of three examples from the MSABTEC Treebank in the CATiB format in which they were annotated.
We  also provide, for comparison,   the analysis  in the increasingly popular Universal Dependency representation \cite{NIVRE16.348,taji2017universal}.

\section{Annotation Process}

The annotation process we followed in the preparation of this treebank is the same process described by \newcite{habash2009catib},
which consisted of the following steps:
{\it (a)  Automatic Tokenization and POS Tagging}, 
{\it (b) Manual Tokenization Correction}, 
{\it (c) Automatic Parsing}, and 
{\it (d) Manual Annotation}.  In this section, we discuss what we did for these steps as well as report on annotator(s), speed and inter-annotator agreement.

\subsection{Annotator(s)}
Due to the relatively small size of our treebank, we had only one annotator working on the task. Our annotator is an educated native Arabic speaker, who was trained on the CATiB scheme and the use of TrEd as part of her work on the original CATiB project \newcite{habash2009catib}.  To evaluate inter-annotator agreement, we worked with a second annotator who was asked to annotate a small part of the treebank (see below).

\subsection{Automatic Tokenization and POS Tagging}
We used MADAMIRA \cite{pasha2014madamira} to tokenize and POS tag the input sentences. We used MADAMIRA's configuration for PATB tokenization and CATiB POS tags.

\subsection{Manual Tokenization Correction}
Our annotator then manually checked and fixed all of the tokenization errors. 
This also included the correction of typos and spelling changes resulting from wrong automatic analysis.
Overall there were 2.8\% tokenization errors, which is higher that MADAMIRA's reported tokenization error rate (around 1.1\%).
The increase is most likely due to the difference in genre between the data used to train MADAMIRA and our corpus.
 
 \subsection{Automatic Parsing} 
 We ran the data with the fixed tokenization through the CamelParser \cite{shahrourcamelparser}, which is trained on the gold CATiB representation of the training data from the PATB parts 1, 2, and 3 according to the splits proposed by \newcite{diab2013ldc}.  We present automatic parsing quality results in Section~\ref{autoparse} %.
 
 \subsection{Manual Annotations}
  The output of the automatic parsing was given in TrEd's \emph{.fs} %NYH not \emph{fx} %DT No, TrEd is .fs
 format to the annotator to manually fix the POS tags, the relation labels, and the syntactic structures of the trees. 

\subsection{Annotation Speed} The manual fixing of the tokenization took the annotator 10 hours of work at the speed of 1,593 words/hour. The manual correction of the parsed trees (POS, relations, and structure) took 40 hours of work at the speed of 466 tokens/hour (398 words/hour). This number is comparable to the speed reported by  \newcite{habash2009catib} (540 tokens/hour).  The sentences in their treebank were of the same genre as the data used to train the automatic parsers unlike our case; furthermore,  their sentences are much longer than ours  (32.0 words/sentence compared to our 7.9 words/sentence). These two issues may explain part of the difference in speed.
The end-to-end speed (from raw words to fully corrected trees) is 319 words/hour.
% NYH I estimated 45 hours.
%remember to plug in the hours for the annotation after you get them from Jamila

\subsection{Inter-Annotator Agreement}
To check the consistency of our annotations, we had another person with previous experience in dependency  annotation annotate a subset of 100 sentences from this treebank. The second annotator started from the CamelParser output on the same corrected tokenization produced by the first treebank annotator. The inter-annotator agreement scores are 98.7\% on POS agreement, 96.1\% on label agreement, 90.6\% on attachment agreement, and 89.7\% on labeled attachment agreement. This is close to the highest average pairwise inter-annotator agreement number reported on the creation of the CATiB Treebank \cite{habash2009catib}.

\section{Results}

We present next a comparison between our treebank and the Penn Arabic Treebank, followed by benchmark results of the performance of a state-of-the-art parser on our corpus.

\subsection{Comparison with Penn Arabic Treebank}
Our  corpus   is from the travel genre, which has some characteristics that are different from those of the news genre. For example, the average sentence length in MSABTEC is 9.31 tokens per sentence, as opposed to PATB's average of 37.57 tokens per sentence. Over 40\% of MSABTEC sentences contained a question, while in PATB this percentage did not exceed 2.6\%. This is expected as travel corpora are more likely to include questions and answers by travellers. 

%more interested in allowing their user to pose questions. %Do we need this justification?  if yes, is this enough, or do we need more?

Moreover, the most frequent words in both corpora vary distinctly. MSABTEC's most frequent verb is <ymkn> \emph{yumkin} `can', which is often used when asking for help. In PATB, however, the most common verb is <qAl> \emph{qAl} `said', which is commonly used for reporting news. In addition, question words such as <km> \emph{kam} `how much', <hl> \emph{hal} `do/does', and <Ayn> \emph{{\AHAMZAUP}yn} `where' appear in the set of the most frequent 50 words in MSABTEC, whereas no question words appear in the set respective to PATB. Frequent nouns in MSABTEC include  <f.dl> \emph{fa{\DAD}l} `favor/please', <rqm> \emph{raqam} `number', and <.grfT> \emph{{\GAYN}urfa{\TAMARBUTA}} `room'. In PATB, the most frequent nouns include <r'iys> \emph{ra{\YHAMZA}iys} `president',
%and <Alr'iys> \emph{Alra{\YHAMZA}iys} `the president', as well as 
<lbnAn> \emph{lubnAn} `Lebanon', <Alywm> \emph{Alyawm} `today', and <Almt.hdT> \emph{Almut{\SHADDA}aHida{\TAMARBUTA}} `the united'.

Another phenomenon that differentiates MSABTEC and PATB is the pronoun frequencies. On the one hand, the most frequent pronouns appearing in MSABTEC are <k>+ \emph{+k}, which is the second person singular pronoun in accusative, and <y>+ \emph{+y} and <ny>+ \emph{+ny}, which are the first person singular pronouns in genitive and accusative case, respectively. On the other hand, the most frequent pronouns appearing in PATB are <h>+ \emph{+h} and <hA>+ \emph{+hA}, which are the masculine and feminine third person singular pronouns, respectively. This leads to the obvious conclusion that MSABTEC mostly contains conversational text that refer to the speaker or the listener, whereas PATB's most dominant style is that of reporting in the third person, which is expected of a news genre corpus. 

\subsection{Automatic Parsing Quality}
\label{autoparse}

We parsed our corpus using CamelParser \cite{shahrourcamelparser}, which was itself trained and optimized on the PATB. Table~\ref{parsing-results} shows the difference in the parser's performance on PATB data, on which it is trained, versus on MSABTEC data. For the PATB, we report on the test set used by \newcite{shahrourcamelparser}. The evaluation of the parser was done using the gold annotations of the MSABTEC data.

\begin{table}[!h]
\centering
\label{parsing-results}
\begin{tabular}{l|l|l|l|}
\cline{2-4}
                           & \textbf{LAS}    & \textbf{UAS}    & \textbf{Label}  \\ \hline
\multicolumn{1}{|l|}{\textbf{PATB}} & 83.8\% & 86.4\% & 93.2\% \\ \hline
\multicolumn{1}{|l|}{\textbf{MSABTEC}} & 73.5\% & 77.0\% & 90.5\% \\ \hline
\end{tabular}
\caption{The evaluation of the CamelParser prediction on data from PATB and MSABTEC}
\end{table}

The error increase in  the results of MSABTEC  from the results of PATB  for the Labeled Attachment Score (LAS), Unlabeled Attachment Score (UAS), and Label selection is 64\%, 70\% and 39\%, respectively. 
%The drop in performance between the results of PATB and the results of MSABTEC for the Labeled Attachment Score (LAS), Unlabeled Attachment Score (UAS), and Label selection is 10.3\%, 9.4\% and 2.7\% absolute, respectively. 
%
This shows that the genre difference between the training data and the testing data significantly affects the performance of the parser. The previously described characteristics that differ between PATB and MSABTEC (sentence length, prevailing person, and different frequent words) can explain this decline in performance. The large performance drop highlights the need for creating treebanks in less-studied genres to support research on them.  %Is this more suitable for the conclusion?

\section{Conclusion and Future Work}
We presented a small dependency treebank of travel domain sentences in Modern Standard Arabic. 
The text comes from a translation of the English equivalent sentences in the Basic Traveling Expressions Corpus. 
The treebank dependency representation is in the style of the Columbia Arabic Treebank. 
Our parsing evaluation of the constructed treebank confirms the need for more treebanks in different genres and domains to 
support research on multi-domain, multi-genre parsers.\\

%\paragraph{}  
In the future, we plan to expand our annotation efforts to other genres and domains as well as to other Arabic dialects. We are also very interested in using the created corpus in improving Arabic syntactic parsing. Since the data we created is small in size compared to the large dominant treebanks, we plan to pursue the genre and domain adaptation research direction.  We also plan to make this resource publicly available to support research on Arabic syntactic parsing.

\newpage

% \nocite{*}
\section{Bibliographical References}
\label{main:ref}

\bibliographystyle{lrec}
\bibliography{ALLBIB-2.5,ExtraBIB}

%\section{Language Resource References}
%\label{lr:ref}
%\bibliographystylelanguageresource{lrec}
%\bibliographylanguageresource{xample}

\end{document}